# AUTOMATING STYLE ANALYSIS AND VISUALIZATION WITH EXPLAINABLE AI - CASE STUDIES ON BRAND RECOGNITION

Yu-hsuan Chen, Levent Burak Kara, Jonathan Cagan

Department of Mechanical Engineering
Carnegie Mellon University
Pittsburgh, PA, 15213, USA

## ABSTRACT

*Incorporating style-related objectives into shape design has been centrally important to maximize product appeal. However, stylistic features such as aesthetics and semantic attributes are hard to codify even for experts. As such, algorithmic style capture and reuse have not fully benefited from automated data-driven methodologies due to the challenging nature of design describability. This paper proposes an AI-driven method to fully automate the discovery of brand-related features. Our approach introduces BIGNet, a two-tier Brand Identification Graph Neural Network (GNN) to classify and analyze scalar vector graphics (SVG). First, to tackle the scarcity of vectorized product images, this research proposes two data acquisition workflows: parametric modeling from small curve-based datasets, and vectorization from large pixel-based datasets. Secondly, this study constructs a novel hierarchical GNN architecture to learn from both SVG's curve-level and chunk-level parameters. In the first case study, BIGNet not only classifies phone brands but also captures brand-related features across multiple scales, such as the location of the lens, the height-width ratio, and the screen-frame gap, as confirmed by AI evaluation. In the second study, this paper showcases the generalizability of BIGNet learning from a vectorized car image dataset and validates the consistency and robustness of its predictions given four scenarios. The results match the difference commonly observed in luxury vs. economy brands in the automobile market. Finally, this paper also visualizes the activation maps generated from a convolutional neural network and shows BIGNet's advantage of being a more human-friendly, explainable, and explicit style-capturing agent.* Code and datasets are available on Github for both phone and car case studies.

## 1. INTRODUCTION

Recognizing, codifying and incorporating desired stylistic objectives into shape design has long been a focus of product development [1,2]. While market appeal is important, conveying specific aesthetic styles through design can be challenging and unpredictable due to the need for differentiation from previous designs and the time-variant nature of style [3,4].

Attempting to relate aesthetic attributes to consumer response and market success, Liu et al., [5] studied three aspects of car aesthetics impact on the market: segment prototypicality (SP), brand consistency (BC), and cross-segment mimicking (CSM). Among all, BC has shown to have the most consistent effect that positively relates to profit, indicating that BC maintenance is one of the key factors of a successful design. To maintain BC, brand feature encoding is crucial for designers to manage the brands' essence and to produce consistent and competitive designs. However, codifying and modeling BC-related features is challenging and subjective, because they are articulated primarily by humans. Because brand features are often subtle and difficult to systematically quantify, designers often have to go through a laborious process to master the brand features of a product.

Previously, research has shown the possibility to construct shape grammars – a sequential and systematic shape description system for product design [6] – to capture product brand. While previous research showed that shape grammars can describe a variety of products' brand features or semantic languages [7–14], the process of finding shape grammars was mostly achieved through human perception, which is a time-consuming and hard to transfer process. Despite the difficulty of automatic shape grammars induction, studies in this field showed the feasibility of constructing describable and quantifiable systems for brand consistency. During such a process, human designers learn the unique features shared among products in each brand. As a realization process resembles supervised learning, a natural question arises: can an AI agent learn brand consistency through a fully automated process and free humans from laborious shape-to-shape comparison? This paper, therefore, models the brand recognition process as a data-driven fine-grained classification task. By examining the trained neural network classifier, humans are expected to gain brand-related feature knowledge from the AI's attention.

There are multiple challenges. First is data scarcity. As most brands only annually release several products that share similar functionality and have distinct exteriors, a class would have only on order of tens of data samples. Since deep learning models' performance hugely relies on a large number of samples to learn meaningful content [15], augmentation techniques to expand the dataset are necessary to study. Second, ensuring the interpretability of the constructed AI imposes great challenges. Since AlexNet [16], convolutional neural networks (CNN) have seen a dramatic accuracy improvement on classifying pixel images. However, extracting key primitives in a parametric fashion from such CNN remains difficult. While class activation mapping (CAM) [17] techniques attempted to interpret CNN pictorially, the resulting heat maps are often fuzzy and lack of precision. Since humans tend to learn, reason, and design based on curves and shapes, this type of architecture and workflow may not effectively capture intuitive brand features. Much less can it be expected to quantify or edit these features using this approach. This paper proposes Brand Identification Graph Neural Network (BIGNet), a curve-based AI that can capture and visualize explicit features. Using the proposed approach, humans can utilize AI as a communicative and explainable style discovery agent and accelerate the design process.

This research's main contributions are:
1. Two vectorized data acquisition approaches for style recognition: parametric modeling from small curve-based datasets, and vectorization from large pixel-based datasets.
2. BIGNet, a novel hierarchical Graph neural network (GNN) that can learn from both scalar vector graphics' (SVG) curve-level and chunk-level features.
3. Evaluation study to produce design insight and feature visualization, which shows BIGNet's capability of perceiving explicit and explainable brand-related features.

## 2. RELATED WORK

This research aims to accelerate product style design, based on deep learning. This section will first review how previous research attempted to construct systematic approaches to convey stylized ideation. Second, this section will review the progress and limitations in fine-grained image classification and curve-based deep learning methods.

*Shape Grammars.* Shape grammars have been used as a computational tool for explicit feature representation and generation for over five decades. A shape grammar consists of a set of shape rules that sequentially eliminate, edit, or generate design primitives [18]. Because of a shape grammar's explicit expression to describe stylized concepts, it later became a feasible method for product designers to capture brand-related features of exterior design [5], which is a crucial subset of branding [19]. Agarwal and Cagan [7] brought shape grammars into industrial products, and they used it to successfully describe coffee machines' shape generation rules and find brands' discriminative features. It was further shown that shape grammars could describe a variety of products' brand features or semantic languages [8–14]. However, the process of defining shape grammars was mostly done by human perception, which is still a time-consuming process and difficult to transfer from one product to another. This research takes a different approach on identifying differentiating geometric features as a means to automate the feature perception task in a data-driven method, thus accelerating the design cycle.

*Fine-grained classification and attention visualization.* Detecting style-oriented features for better object recognition accuracy has been a challenge. However. explicit detectors [20,21] and descriptor design [22–24] paved the way for deep CNNs [25] to reach high accuracy by learning complex and transformation-invariant features [16,26]. To generalize CNN's application to a variety of tasks, finetuning on pre-trained networks by only training one fully connected layer from scratch was studied and found to result in much faster convergence and better accuracy [27–29]. These advancements, however, couldn't promise CNN to have full interpretability of the describable features. While it showed possibility to visualize localized regions on fine-grained classification [17,30–33], because CNN is learning from pixelated information, it is still unclear which shapes or curves are important, therefore diminishing the usage for designers. Therefore, the attempt to train a deep learning model on curve-based image representation is proposed in this work to visualize human-readable features.

*Curve-based recognition methods.* As representing images in curves is much closer to how humans see an image, research on curve-based recognition models focuses on building AI that can learn descriptive features from human sketches for classification. In the field of sketch recognition, early studies [34–36] use Support Vector Machine (SVM) as the classifier to differentiate rasterized sketch images. To visualize stroke importance, Schneider and Tuytelaars [35] also adopted a leave-one-feature-out (LOFO) technique to remove one stroke at a time and see how it affects the classification score. More recent studies [37–39] took advantage of CNN and achieved better performance and robustness on the recognition task. To recognize curve-based images with multiple abstract levels of features, Yu et al., [37] proposed a multi-scale, multi-channel CNN architecture to learn from partial images segmented from stroke order. However, it still has not incorporated grouping information of curve-based images, which is useful for learning more descriptive features [40]. To address this limitation, Li et al., [38] then proposed sketch-R2CNN to classify sequentially rendered images and paired with a recurrent neural network (RNN) attention mechanism that enabled better accuracy and feature visualization. Although sketch recognition has made significant progress, most studies focused on recognizing simple human sketches comprising only a few tens of strokes. These strokes are typically drawn with straight lines rather than curves, and the sketches are often low resolution, which is a simple representation and limits the complexity that can be achieved with the task. As scaling the recognition onto industrial products like cars can easily have thousands of curves, previous studies would not be applicable due to expensive rendering computation, and vanishing gradient problem for RNN-based architectures. Furthermore, industrial products typically consist of dozens to hundreds of groups of curves (chunks) that may contain higher-level brand-related features. There has been little research on analyzing the inter-chunk relationships at a scale of 10,000 or more. The above studies conclude that since CNN is restricted to work on pixelated images, it is not possible to analyze SVGs without first rasterizing, which results in a sparse image and loses grouping information.

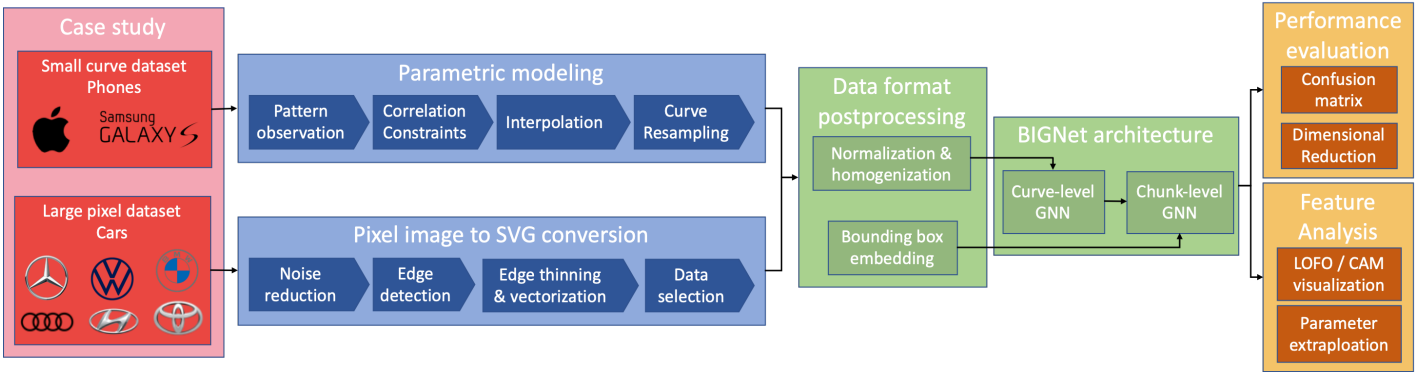

**Figure 1:** Workflow of this research, illustrating the key stages of product, brand, and model selection (red), data acquisition (blue), neural network's message passing design (green), and AI analysis (yellow). The logos of phones and cars are shown to represent the brands selected for identification, but the classification task is based on product shapes rather than logos. Two case studies demonstrate the framework's adaptability to different product domains and scales on the design spectrum.

*Spatial Graph Neural network.* GNN refers to the domain of deep learning methods designed to deduce information from general non-Euclidean graphs. A graph $\mathcal{G}$ is defined as $\mathcal{G}(\mathcal{V}, \mathcal{E})$ where $\mathcal{V}$ is the set of vertices or nodes and $\mathcal{E}$ is the set of edges [41]. Among the branches, spatial-based convolution GNNs (convGNN) are found to be most analogous to conventional CNNs, while allowing the nodes to have an arbitrary number of neighbors and offering flexibility on connectivity strength as well as the aggregation process. Neural Network for Graphs (NN4G) [42] was the first work towards spatial ConvGNNs that performed aggregation by summing up each node's neighborhood information directly. After that, multiple architecture improvements were proposed around flexible aggregation [43–45] and sampling [46] strategies. Among all, Graph Attention Networks (GAT) [45] adopted attention mechanisms to learn the relative energy (weights) between two connected nodes. By enabling specification of different weights to different nodes, it has achieved state-of-the-art prediction results in Cora, Citeseer, and Pubmed benchmarks. From an engineering perspective [47], GNN has shown its capability of tackling various problems including physical modeling [48], chemical reaction prediction [49], traffic state prediction [50], and engineering drawings' segmentation [51,52]. Inspired by the recent successful applications, this research models each SVG as a two-tier graph and builds a spatial GNN with learnable chunk-level attention mechanisms to perform graph-level classification.

## 3. METHODOLOGY

To enhance designers' ability to edit parametric curves in real time and evaluate the impact on brand consistency, the goal of this research is to build a curve-based AI-driven feature retrieval surrogate that is both explainable and describable. As brand consistency is shown to be important yet abstract for humans to easily identify, the case studies of the proposed methodology are applied to industrial products' brand recognition. The research workflow is shown in Figure 1.

### 3.1 Data representation and acquisition

This research focuses on the front view of product models because just as human beings are more recognizable by their faces, designers tend to place the most recognizable features in products' front view as well [53]. SVG format is chosen to represent the objective products, as it composes an image of chunks of geometry defined explicitly by parametric control points. To maintain data homogeneity, all curves are converted to cubic Bezier curves, with each image represented by an arbitrary number of chunks of curves, while each curve is parameterized by four control points (eight scalar values).

Facing the scarcity of product data, this research first synthesizes intermediate designs as a data augmentation method. This is achieved by creating and interpolating unified design rules for each brand of products through human observation. For the simplicity of geometry and planar design, a case study is run to identify mobile phones' exterior features. Second, this research attempts to generate SVGs from vectorizing a generic pixel dataset. As images are taken from the same car at slightly different post angles, this perspective difference contributes to part of the data augmentation. This is implemented with an image processing pipeline including background removal, noise reduction, edge detection and vectorization. A case study on cars' recognition is run for the second approach.

### 3.2 AI architecture

To learn the discrepancy among curve-based images in terms of brand-related styles, this paper proposes Brand Identification Graph Neural Network (BIGNet), a two-tier spatial GNN that can learn from the SVG format dataset (Figure 2). In the first layer, a chunk of curves is represented as a graph, while each node is a curve, and connectivity is determined by its neighbor curves. More precisely, the model first samples and aggregates the neighborhood of each node, feeds each node into fully connected layers (FCs), and then reads out the response by average pooling. In the second layer, an SVG picture is represented as a graph, while each node is a chunk of curves, and connectivity strength is determined via the weights learned from the bounding box parameters. After aggregation, the hidden layers are concatenated and passed to the last fully connected layer to get the prediction. BIGNet's forward propagation is summarized in Algorithm 1. The parameters used in the two case studies are slightly different to adjust for the images' complexity level and are listed in Table 1.

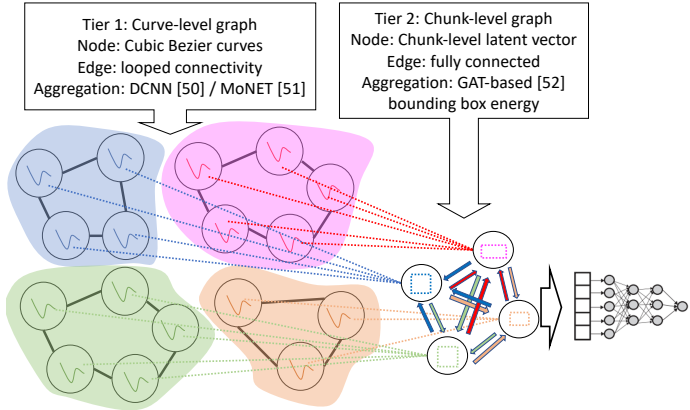

**Figure 2:** Schematic diagram of the two-tier BIGNet structure.

### 3.3 AI evaluation overview

After successfully training the network, a series of evaluation criteria is studied to deduce explainable and quantifiable results.

The evaluation of the synthetic phone dataset aims to determine whether BIGNet can accurately perceive intricate details, and whether these details can also be transferred to humans. As the dataset is synthesized using shape rules created by human through observation from a reference source dataset, the accuracy of the reference source dataset is first examined to verify the manually constructed shape rules' validity. After that, a dimensional reduction is performed on the networks' latent vector to see if the brands are well separated. To further investigate the network's attention, an ablation study using leave-one-feature-out (LOFO) is performed both at the curve and chunk level. This test removes a chunk or a curve from the picture at a time. Those curves that result in a prediction performance drop when removed are considered to be important features. Finally, based on the localized features observed from the ablation study, parameter extrapolation (Partial Dependence Plot [54]) is implemented on the original shape rules to visualize the confidence change. This step further checks the importance of highlighted curves to the discrimination task to understand if the products' brand features are successfully extracted by the AI.

The evaluation of BIGNet on the vectorized car dataset assesses its ability to recognize brand-related styles in complex, automatically generated data. This includes testing the model's

**Algorithm 1:** BIGNet forward propagation algorithm

**Input:** SVG graph $\mathcal{G}(\mathcal{V}_1, \mathcal{E}_1)$; graph-level FCs $f_1$; aggregated chunk-level FCs $f_2$; primitive chunk-level FCs $f_3$; curve-level FCs $f_4$; chunk attention matrix FC $f^*$; chunk-level depth $D_1$; curve-level depth $D_2$; chunk attention matrix $W_1$; curve diffusion weight $W_2$; chunk aggregating function $A_1$; curve aggregating function $A_2$; chunk graphs $\{\mathcal{G}(\mathcal{V}_2(v_1), \mathcal{E}_2(v_1)), \forall v_1 \in \mathcal{V}_1\}$; curve features $\{\{x_{v_2}, \forall v_2 \in \mathcal{V}_2(v_1)\}, \forall v_1 \in \mathcal{V}_1\}$

**Output:** Vector representation $z_\mathcal{G}$

for $v_1 \in \mathcal{V}_1$ do
$\quad$for $v_2 \in \mathcal{V}_2(v_1)$ do
$\quad\quad h_{v_2}^0 \leftarrow x_{v_2}$
$\quad\quad e_{v_2}^0 \leftarrow x_{v_2}$
$\quad\quad$for $d_2 = 1 \ldots D_2$ do
$\quad\quad\quad e_{v_2}^{d_2} \leftarrow A_2(e_{v_2}^{d_2-1}, \mathcal{E}_2(v_2), W_2)$
$\quad\quad\quad h_{v_2}^{d_2} \leftarrow concat(h_{v_2}^{d_2-1}, e_{v_2}^{d_2})$
$\quad\quad$end
$\quad$end
$\quad H_{v_1}^* \leftarrow pool(\{f_4(h_{v_2}^{D_2}), \forall v_2 \in \mathcal{V}_2(v_1)\})$
$\quad H_{v_1}^0 \leftarrow f_3(H_{v_1}^*)$
$\quad E_{v_1}^0 \leftarrow H_{v_1}^0$
$\quad$for $d_1 = 1 \ldots D_1$ do
$\quad\quad E_{v_1}^{d_1} \leftarrow A_1(E_{v_1}^{d_1-1}, \mathcal{E}_1, f^*(W_1))$
$\quad\quad H_{v_1}^{d_1} \leftarrow concat(H_{v_1}^{d_1-1}, E_{v_1}^{d_1})$
$\quad$end
end
$H_\mathcal{G}^* \leftarrow pool(\{f_2(H_{v_1}^{d_1}), \forall v_1 \in \mathcal{V}_1\})$
$z_\mathcal{G} \leftarrow f_1(H_\mathcal{G}^*)$

**Table 1:** BIGNet's parameters for the two case studies.

| Case | Phone | Cars |
|---|---|---|
| Activation | LeakyReLU | |
| Optimizer | Adam | |
| Pooling | Average pooling | |
| Loss | Binary cross entropy | Categorical cross entropy |
| $\mathcal{E}_2$ | Loop graph (each node has 2 neighbors) | |
| $D_2$ | Bidirectional, depth: 2 | Bidirectional, depth: 2 |
| $A_2$ | $e_{v_2}^{d_2} = (1-W_2)e_{v_2}^{d_2-1}$ $+W_2 \times \mathcal{E}_2 e_{v_2}^{d_2-1}$ | $e_{v_2}^{d_2} = (1-W_2)e_{v_2}^{d_2-1}$ $+W_2 \times Linear(8 \to 8)(\mathcal{E}_2 e_{v_2}^{d_2-1})$ |
| $W_2$ | 1 | 0.5 |
| $f_4$ | $Linear(32 \to 24 \to 12)$ | $Linear(24 \to 32 \to 24)$ |
| $f_3$ | pass | $Linear(24 \to 24 \to 24)$ |
| $\mathcal{E}_1$ | Fully connected | |
| $D_1$ | 2 | 2 |
| $W_1$ | $N \times 5$ | $N^2 \times 5$ |
| $f^*$ | $Linear(5 \to 12)$ | $Linear(5 \to 24)$ |
| $A_1$ | $E_{v_1}^{d_1} = f^*(W_1) \times E_{v_1}^{d_1-1}$ | |
| $f_2$ | pass | $Linear(72 \to 24 \to 24 \to 24)$ |
| $f_1$ | $Linear(36 \to 18 \to 8 \to 2)$ | $Linear(24 \to 18 \to 12 \to brand\#)$ |
| Learnable parameters | 2000 ~ 2076 | 6716 ~ 6812 |

robustness and consistency across different tasks and scenarios. Firstly, Confusion matrix and dimension-reduced latent vectors plots are calculated to learn the brands' differences in distinguishability. Secondly, AI's chunk-level attention is visualized using a class activation mapping (CAM) [17]–inspired algorithm. By highlighting the chunk that contributes more to correct identification, CAM is shown to have much more robustness than LOFO on SVGs with higher order of chunks. Finally, to compare the curve-based approach to the pixel-based approach, a CNN is finetuned using ResNet-50 that was pre-trained with simCLR [28]. The attention of this CNN is then visualized using Grad-CAM [33]. As CNN is expected to reach a better accuracy level due to its quantitively much larger model size, this study focuses on examining whether BIGNet conveys more explicit and describable design features than CNN's class attention visualization.

## 4. CASE STUDY - PHONES
### 4.1 Brand and model selection

This study compares and differentiates the front views of the products of the two most popular cellphone brands – Apple and Samsung [55,56]. Among the many lines of Samsung phones, the Samsung Galaxy S series has the most similar functionality and price range as Apple's iPhone and therefore is chosen to be the competitor of Apple's iPhone. To preserve a reasonable degree of homogeneity, all the phone models chosen are without home buttons, which are also more contemporary designs.

### 4.2 Parametric modeling from small curve datasets
#### 4.2.1 Synthetic Dataset Generation

Challenges exist in finding an abundant and well-measured dataset. After realizing the need of increasing the sample size of an existing dataset collected from Dimensions.com (shown in Table 2a), this study then observes the patterns for the selected models, and by using parameter interpolation on the manually established shape rules (number and types of parameters are listed in Table 2b, and shape rules example of Apple is shown in Figure 3a), a synthetic SVG dataset with 20,000 synthetic phones, 10,000 for both Apple and Samsung is successfully created (some results are shown in Figure 3b).

#### 4.2.2 Synthetic image preprocessing

After generating the synthetic SVG dataset, the next important step is to adapt it to a more homogenous format, so that the AI can learn the difference between brands' shapes instead of the difference between the brands' creative processes. Therefore, this study then rasterizes the images and vectorizes each into a cubic Bezier SVG using Potrace [57]. All the phones' heights are then normalized to 1 since the synthetic dataset has relatively larger Samsung phones than Apple phones. This will enable the AI to learn meaningful design languages.

### 4.3 Results and discussion of phone case study
#### 4.3.1 Model's training process and performance

Using BIGNet's architecture and parameters from Table 1 column 1, after 105 epochs, the model was able to reach over 99.8% for both training and testing accuracy (Figure 4) and can 100% predict the reference sources' brands as well (Table 3a). The small difference between train and test accuracy could be attributed to the fact that both sets are generated using identical interpolated shape rules. Moreover, the high accuracy result is likely due to the same reason which gives the two classes significant brand consistency, and therefore makes the problem highly manageable for BIGNet. Table 3b shows the confusion matrices on the three datasets.

**Table 2a:** The phone models of an existing dataset.

| Apple iPhone | Submodel Names | Total Number of Submodels |
|---|---|---|
| X | XR, X, XS, XS Max | 15 |
| 11 | 11, 11 Pro, 11 Pro Max | |
| 12 | 12 mini, 12, 12 Pro, 12 Pro Max | |
| 13 | 13 mini, 13, 13 Pro, 13 Pro Max | |

| Samsung Galaxy S | Submodel Names | Total Number of Submodels |
|---|---|---|
| S10 | S10e, S10, S10+, S10 5G | 7 |
| S20 | S20, S20+, S20 Ultra | |

**Table 2b:** Number and types of parameters used in shape rules to make the synthetic SVG dataset.

| Shape Rules Parameters | Continuous (ex: height, width, fillet) | Discrete (ex: lens position) | Regulation (ex: height-width ratio) |
|---|---|---|---|
| Apple | 28 | 5 | 6 |
| Samsung | 25 | 1 | 12 |

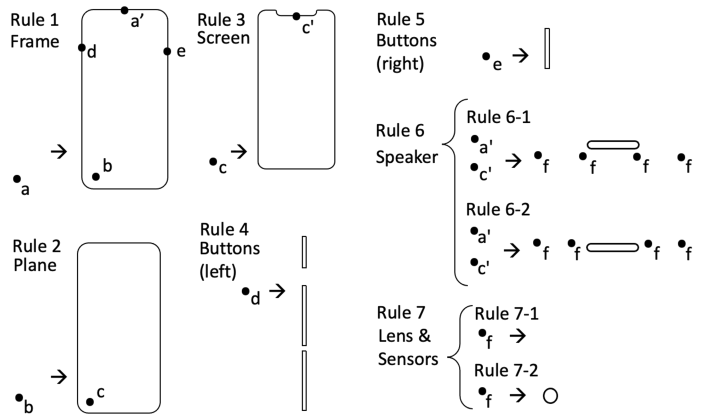

**Figure 3a:** Shape rules used to generate Apple phones. By sequentially applying rule 1 to rule 7 with parameters of Apple in table 2b, a synthetic Apple phone is made.

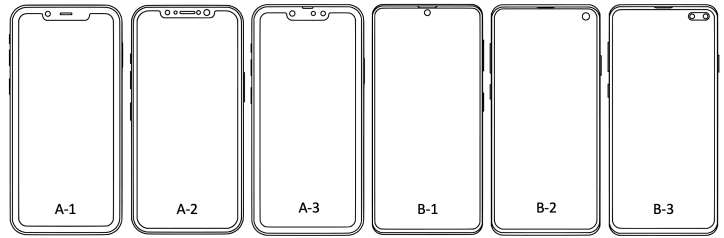

**Figure 3b:** Synthetic examples of Apple (A-1~A-3) and Samsung (B-1~B-3). For Apple, there are two distinct types of speaker position: at the middle (A-1, A-2) or at the top (A-3) of the notch. The number of circles representing lens and sensors range from zero to four. For Samsung, there are three distinct types of lens: one at the middle (B-1), one at the upper right corner (B-2) and two at the upper right corner (B-3).

### 4.3.2 Dimensional reduction

By visualizing the last hidden layer with t-distributed Stochastic Neighbor Embedding (t-SNE) and Principal Component Analysis (PCA) in Figure 5, Apple phones (red) and Samsung phones (blue) are clustered and well-separated from each other, demonstrating that the network can very clearly discriminate the two phone brands.

### 4.3.3 LOFO visualization study

To visualize localized features, important chunks are colored red, and curves are colored blue for each picture (some results are shown in Figure 6). Discriminative features of the two brands are then observed and summarized in Table 4. Since both brands highlight lens, fillet, width and the gaps between screen and the frame, the study then examines partial dependence plots created by parameter extrapolation on these features in the following section to further understand the network's attention.

### 4.3.4 Partial dependence plot

(1) Lens horizontal position

The goal of this experiment is to see if prediction confidence drops while extrapolating Apple's lens horizontal position. Since the trained BIGNet is robust by looking at multiple features, the result in Figure 7 is plotted when feature i-2, i-4 and i-6 are all shifted to Samsung's dimensions range. The confidence curve drops when the lens is at the middle and to the right of the notch, which is because those are both Samsung's possible lens locations (see b-1, b-2 in Figure 6).

(2) Fillet Radius and height-width ratio

Similar results from (1) are found while extrapolating Apple's width. The interesting thing is although Samsung is shorter in dimension of both width and normalized width, the model considers wider phones as Samsung. This is the result from the model looking at the length of the frame's segment instead of the whole width, which also takes fillet radius into consideration (Table 5). In Figure 8, this explanation is verified since the crossover width lays between the boundary of the two brands' normalized segment length ranges.

(3) Screen-frame gaps

Since both brands also highlight the gaps among screen, frames' inner width (plane) and outer width (edge) in 4.3.3, a 2-D extrapolation experiment on the two gap parameters is done on Samsung's shape rules. In Figure 9, results show that phones with smaller gaps between the screen and the frame are more likely to be predicted as Samsung, which matches the interpolation range of the two brands.

**Table 3a:** Trained model's loss and accuracy on the 3 datasets.

| Dataset | number of samples | loss | accuracy |
|---|---|---|---|
| train | 18000 | 0.0053 | 99.83% |
| test | 2000 | 0.0038 | 99.85% |
| reference | 28 | 0.0046 | 100% |

**Table 3b:** Confusion matrices (not normalized)

| Train Set Confusion Matrix | | |
|---|---|---|
| truth \ prediction | Apple | Samsung |
| Apple | 9000 | 0 |
| samsung | 30 | 8970 |

| Test Set Confusion Matrix | | |
|---|---|---|
| truth \ prediction | Apple | Samsung |
| Apple | 999 | 1 |
| samsung | 2 | 998 |

| Reference Set Confusion Matrix | | |
|---|---|---|
| truth \ prediction | Apple | Samsung |
| Apple | 15 | 0 |
| samsung | 0 | 13 |

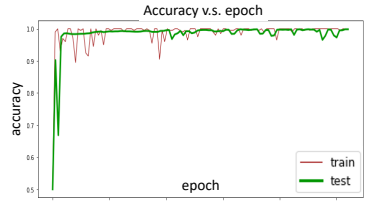

**Figure 4:** Accuracy converges to 98% within 20 epochs.

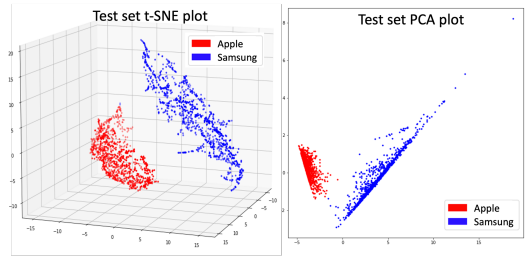

**Figure 5:** t-SNE and PCA plots from test set's latent vectors.

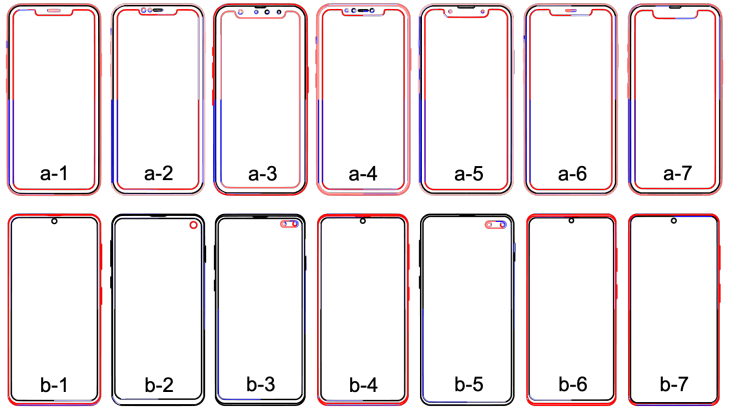

**Figure 6:** LOFO results from Apple (a-1~7) and Samsung (b-1~7).

**Table 4:** A summary of the model's attention by observing the LOFO visualization results. One thing to notice is i-1~4 are similar parameters to s-1~4, therefore this study continues to experiment on their parameter extrapolations.

| Brand | index | Observed Features | Figure |
|---|---|---|---|
| Apple | i-1 | lens | a-2, a-3, a-4, a-5 |
| | i-2 | corner's fillet | a-1, a-3, a-5, a-7 |
| | i-3 | width | a-1, a-2, a-3, a-4, a-6, a-7 |
| | i-4 | screen-frame gap | a-1, a-2, a-3, a-4, a-5, a-6, a-7 |
| | i-5 | speaker when at the middle | a-1, a-2, a-4, a-6 |
| | i-6 | notch related features | a-1, a-2, a-4, a-5, a-6, a-7 |
| | i-7 | mute button | a-1, a-2, a-4, a-5, a-6 |
| Samsung | s-1 | lens when at the right corner | b-2, b-3, b-5 |
| | s-2 | corner's fillet | b-1, b-2, b-3, b-5, b-7 |
| | s-3 | width | b-1, b-2, b-3, b-4, b-5, b-6, b-7 |
| | s-4 | screen-frame gap | b-1, b-2, b-4, b-6, b-7 |

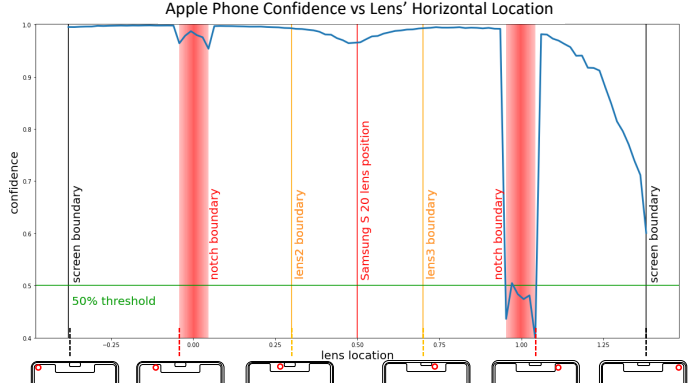

**Figure 7:** Confidence change while extrapolating Apple's lens horizontal position.

**Table 5:** Although Samsung has a shorter normalized width, it also has a relatively smaller fillet radius, therefore its segment length that the model perceives is longer (last column).

| Dimension / Brand | Width (mm) | Fillet (mm) | Height (mm) | Normalized width | Normalized segment length |
|---|---|---|---|---|---|
| Apple | 71.15 | 10.75 | 146.15 | 0.49 | 0.34 |
| Samsung | 73.1 | 7.415 | 154.55 | 0.47 | 0.38 |

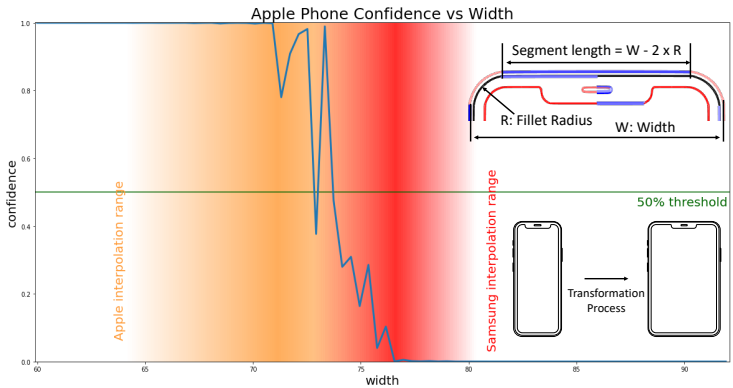

**Figure 8:** Confidence change while extrapolating Apple phone's width. The orange and red regions represent the range of Apple and Samsung's normalized segment length.

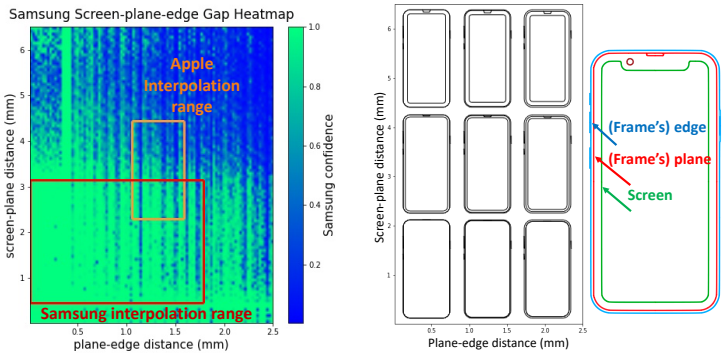

**Figure 9:** Since Samsung has shorter distances between both screen to plane and plane to edge, the heatmap shows a greater prediction confidence at the lower left corner, meaning this is also a discriminative feature to the model.

## 5. CASE STUDY - CARS
### 5.1 Background

While the phone case study has shown the viability of GNN learning from curve-based representation, laborious work has to be done on observing and parameterizing to synthesize an augmented dataset. It is doable on two phone brands promptly, but as unified, interpolatable parametric expressions have to be established for every studied model in every brand; generalizing without human attention is challenging. In addition, for products like cars with more complex shapes and more variety of models, although it will be difficult to construct unified shape grammars, there exist pixel datasets that have thousands of images for each brand. If SVGs can be acquired from such resources, BIGNet can be applied to learn from these large datasets with complex product geometries, and the workflow of extracting brand-related features can be even further automated.

This case study, therefore, aims to explore the feasibility of converting pixel images to curve images to create a data-driven, hands-free recognition system. Since cars have distinctive functionality and design criteria that differ significantly from those of phones, this study not only demonstrates the potential of fully automated SVG retrieval but also attempts to showcase the adaptability and generalizability of BIGNet across different product domains and design scales. Therefore, the following distinct yet comparable training scenarios are run to examine the model's flexibility, robustness, consistency and explainability:

*Classifying different number of brands*. All vectorized images in this case study are generated from the same automated pipeline without the need of parametric modeling. As a result, expanding the number of models and brands for a more comprehensive style classification is made possible with little human effort. Yet, one of the counters of deep learning methods is their lack of reproducibility. This is caused by having redundant freedom of parameters that would lead to suboptimal convergence. However, a style perception agent is expected to consistently exhibit the same features regardless of the training scheme, or data processing nuances, to enable designers ability to reason and make decisions from its inference. To investigate the generalizability of BIGNet and showcase the ease of dataset regeneration, this study conducts both six- and ten-brand classifications. Although a decrease in overall confidence when moving from six to ten brands is expected, it is also anticipated that BIGNet will still exhibit similar patterns in terms of which brands are easily identifiable and which are not.

*Logo removal*. While logos aid in brand recognition, the geometry of logo is not necessarily the brand-related features design engineers are attempting to extract. Therefore, identifying logos as part of the learned brand features is not the primary objective, as this may lead to overfitting on the logo and hinder the attention given to other important design features. To assess the effect of logos on brand classification, separate models are trained on cars with logos and without logos.

*Comparison to CNN trained on pixel images*. BIGNet trained on SVGs is claimed to offer more explainability, but CNN trained on pixel images is widely used and offers high identification accuracy. To compare the two approaches, this study finetunes a simCLR-pretrained ResNet-50 using the exact train-test split of pixel images before vectorization. Since ResNet-50 has 23 million learnable parameters and pixel images contain richer information, CNN is expected to have better accuracy than BIGNet. Despite that, as the goal of this research is to extract explicit and usable attention features, this comparison will focus on comparing feature visualization mapping and validate whether BIGNet provides more explicit and parametric results.

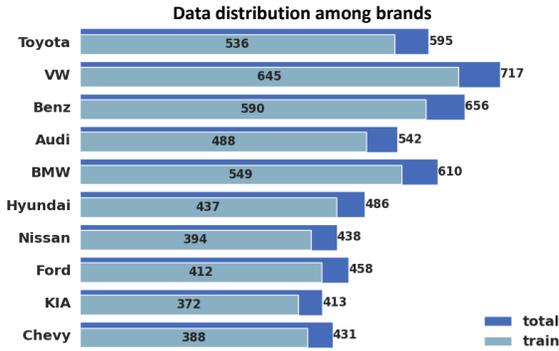

**Figure 10:** Data distribution among brands shows a nonnegligible imbalance. The largest class Volkswagen (VW) has 73.6% more data than the smallest class KIA. Additionally, with only several hundreds of samples per brand, this study performs a 9:1 stratified train-test split.

**Table 6:** Data distribution among segments and years. Among all brands, major segments are sedan, SUV, and hatchback, while most images come from 2009-2015. This demonstrates not only the balanced segment and year ratio in each brand, but also validates the concurrent brand competency during 2009-2015.

| brand / segment | Toyota | VW | Benz | Audi | BMW | Hyundai | Nissan | Ford | KIA | Chevy | total | ratio |
|---|---|---|---|---|---|---|---|---|---|---|---|---|
| sedan | 114 | 190 | 119 | 127 | 231 | 171 | 72 | 76 | 153 | 94 | 1347 | 38% |
| SUV | 108 | 23 | 151 | 59 | 117 | 87 | 57 | 24 | 35 | 32 | 693 | 20% |
| hatchback | 74 | 80 | 27 | 29 | 28 | 40 | 50 | 88 | 63 | 74 | 553 | 16% |
| MPV | 34 | 55 | 50 | 0 | 0 | 10 | 9 | 35 | 65 | 0 | 258 | 7% |
| others | 21 | 166 | 110 | 117 | 97 | 0 | 60 | 38 | 13 | 36 | 658 | 19% |
| labeled % | 59% | 72% | 70% | 61% | 78% | 63% | 57% | 57% | 63% | 55% | 64% | |

| brand / year | Toyota | VW | Benz | Audi | BMW | Hyundai | Nissan | Ford | KIA | Chevy | total | ratio |
|---|---|---|---|---|---|---|---|---|---|---|---|---|
| <2008 | 22 | 21 | 8 | 6 | 8 | 29 | 28 | 8 | 11 | 0 | 141 | 3% |
| 2008 | 6 | 27 | 6 | 4 | 31 | 22 | 22 | 36 | 15 | 7 | 176 | 3% |
| 2009 | 53 | 72 | 34 | 36 | 50 | 61 | 17 | 40 | 25 | 26 | 414 | 8% |
| 2010 | 58 | 64 | 81 | 102 | 61 | 37 | 51 | 16 | 31 | 70 | 571 | 11% |
| 2011 | 61 | 105 | 75 | 33 | 108 | 78 | 54 | 47 | 67 | 60 | 688 | 13% |
| 2012 | 111 | 122 | 80 | 74 | 63 | 44 | 61 | 83 | 64 | 48 | 750 | 14% |
| 2013 | 110 | 112 | 116 | 129 | 123 | 81 | 91 | 151 | 56 | 86 | 1055 | 20% |
| 2014 | 127 | 121 | 155 | 107 | 129 | 78 | 87 | 31 | 111 | 100 | 1046 | 20% |
| 2015 | 39 | 72 | 96 | 46 | 31 | 51 | 21 | 42 | 33 | 32 | 463 | 9% |
| >2015 | 4 | 1 | 2 | 5 | 6 | 5 | 1 | 3 | 0 | 2 | 29 | 1% |
| labeled % | 99.3% | 100.0% | 99.5% | 100.0% | 100.0% | 100.0% | 98.9% | 99.8% | 100.0% | 99.8% | | |

### 5.2 Data selection and preprocessing

HK Comp cars [58], the largest brand- and orientation-labeled car dataset, is chosen to be the source of pixel images. To ensure a fair representation of the market, this research selects the top six and top ten most abundant car brands as the dataset, including both luxurious and affordable brands with an international distribution in Asia, Europe, and America. While the resulting selection is expected to be non-biased and diverse, the total number of images per brand is relatively small, with only a few hundred images per brand. Therefore, to provide enough training data for BIGNet, a train-test ratio of 9:1 is split. Nonetheless, the dataset has an imbalanced number of images among brands (Figure 10). To address this issue, two steps are taken to ensure no falsely high accuracy will occur due to bias towards brands with more data. First, a stratified split is performed to preserve the same ratio in both sets. Second, the algorithm randomly oversamples minority classes during training to ensure each brand has equal representation. More statistics on data distribution, including year and car segments, are provided in Table 6 to ensure the variance of each brand.

To achieve successful vectorization, an image preprocessing pipeline is proposed. First of all, background noise is removed by applying detectron2 [59], a maskRCNN-based AI to detect and apply the mask on the original image. Second, Google cloud vision API is used to detect and remove the logo for a comparison dataset. Edge detection is then implemented before curve-fitting. Through comparing state-of-the-art edge detection methods, the transformer-based EDTER [60] is found to achieve the best performance at preserving object curves and eliminating reflections on the cars' glossy surfaces. After applying edge thinning on EDTER's response, Potrace [57] is used to vectorize the edges into SVG. To maintain a reasonable degree of homogeneity, all vectorized results are converted to cubic Bezier curves. At last, each SVG has its height normalized to 1, and the bounding box information of center coordinate, width, height, and area are pre-computed to enhance chunk-level aggregation.

### 5.3 AI modification for increased data complexity

Since BIGNet is trained with the same type of image format, the message passing flow in this case study shares a lot of common blocks in the phone study's architecture. However, cars' exterior shape has many more components than phones, and the vectorization from generic images also unavoidably leads to redundant curves. The two factors result in around 20 times more chunks and 10 times more curves than phones. To cope with the data's increased scale of complexity, BIGNet architecture is modified from the phone case study's parameters (Table 1 column 2), and can be summarized into three aspects:

(1) *Increase the size of GNN*. First, a more flexible curve-level aggregation policy with one fully connected layer (FC) is enhanced. Second, chunk-level FCs are added for better digestion both before and after the chunk-level aggregation. Lastly, the widths of most FCs are doubled to allow GNN's bandwidth of carrying more features per node. This is a reasonable modification because the geometry of car parts is much more complicated than that of a phone's, with a lot more organic shapes. Overall, number of learnable parameters is increased from 2000 to 6716.

(2) *Better use of bounding box information*. One of BIGNet's key components is the chunk-level connectivity strength learned from bounding box attention. In the phone study, as all data share the common largest chunk being the phone's outer frame, the connectivity strength matrix is derived from the bounding box relationship normalized by the maximum bounding box. Such homogeneity doesn't exist on vectorized

SVGs of cars. Furthermore, inter-chunk relationship, which is the square of number of chunks, becomes roughly 100 times larger than synthetic phones, and has a much larger variance. All these factors impose great challenges to the previous parameters. To tackle the increased complexity, the ratio values in the correlation matrix, namely area, width and height, are first normalized to between 0 and 1 by taking logarithmic values. After that, an FC is applied to adapt the shape of desired features. The pseudo-code is showed in Algorithm 2.

(3) *Augmentation*. First, horizontal flip is done because cars' front views are symmetric. Second, another augmentation is applied by running two distinct EDTER models to retrieve slightly different edge detection response. The two combined techniques enlarge the dataset four times. Lastly, in this dataset, multiple images are often taken on the exact make and model at slightly different perspectives of front view. As humans can identify car parts from slightly off perspectives, this research also treats this as a natural perspective augmentation. It is expected to prevent GNN from overfitting and therefore achieve a more robust model.

**Algorithm 2:** chunk level aggregation in car case study

**Input:** graph with N nodes $\mathcal{G}(\mathcal{V} = \{v_1 \ldots v_N\}, \mathcal{E})$;
Node features: $E_{N \times m} = \{e_{1 \times m}(v_1) \ldots e_{1 \times m}(v_N)\}$;
Nodes' bounding box features (horizontal location, vertical location, width, height, area):
$\beta_{1 \times 5}(v) = \{x(v), y(v), w(v), h(v), a(v)\}, \forall v \in \mathcal{V}_N$;
Linear layer $f^* = f^*(5 \to m)$

**Output:** Node features after one aggregation:
$E'_{N \times m} = \{e'_{1 \times m}(v_1) \ldots e'_{1 \times m}(v_N)\}$

Init $B_{primitve} = \mathbf{0}_{N \times N \times 5}$
Init $B_{adapted} = \mathbf{0}_{N \times N \times m}$
Init $E'_{N \times m} = \{e'_{1 \times m}(v_1) \ldots e'_{1 \times m}(v_N)\} = \mathbf{0}_{N \times m}$
Init $\mathbf{E}' = \{\mathbf{e}'_{N \times m}(v_1) \ldots \mathbf{e}'_{N \times m}(v_N)\} = \mathbf{0}_{N \times N \times m}$
**for** $i$ in 1…N **do**
    **for** $j$ in 1…N **do**
        $B_{primitve}[i, j, 0] \leftarrow x(v_i) - x(v_j)$
        $B_{primitve}[i, j, 1] \leftarrow y(v_i) - y(v_j)$
        $B_{primitve}[i, j, 2] \leftarrow \log(\frac{w(v_i)}{w(v_j)})$
        $B_{primitve}[i, j, 3] \leftarrow \log(\frac{h(v_i)}{h(v_j)})$
        $B_{primitve}[i, j, 4] \leftarrow \log(\frac{a(v_i)}{a(v_j)})$
        $B_{adapted}[i, j, :] \leftarrow f^*(B_{primitve}[i, j])$
    **end**
    $\mathbf{e}_{N \times m}(v_i) \leftarrow stack(e_{1 \times m}(v_i))\ N\ times$
    $\mathbf{e}'_{N \times m}(v_i) \leftarrow B_{adapted}[i, :, :] \cdot \mathbf{e}_{N \times m}(v_i)$
    $e'_{1 \times m}(v_i) \leftarrow pool(\mathbf{e}'_{N \times m}(v_i), axis = 0)$
**end**

## 5.4 Results and discussion of car case study

This section will first focus on the results of BIGNet's six-brand classification without logo removal, and then have an extensive comparison with other training scenarios.

### 5.4.1 Training

During training, batch size is chosen to be 100, and learning rate is initialized to be 0.001. After ~27000 iterations, the learning rate is decreased to 0.0001 for fine-tuning since both train and test accuracy are stagnated (Figure 11). As the end of the model starts overfitting, maximized test accuracy, which is at the 723rd epoch, is selected for evaluation. It is able to reach 89.3% training accuracy and 80.6% test accuracy.

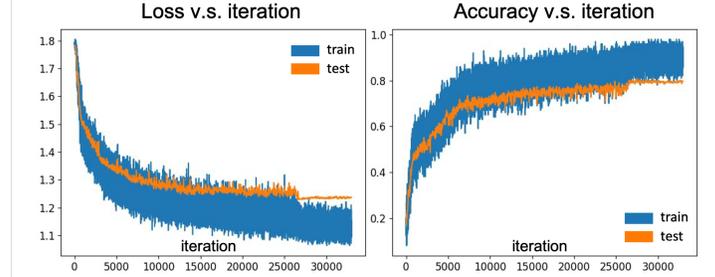

**Figure 11:** Accuracy and loss during the training process.

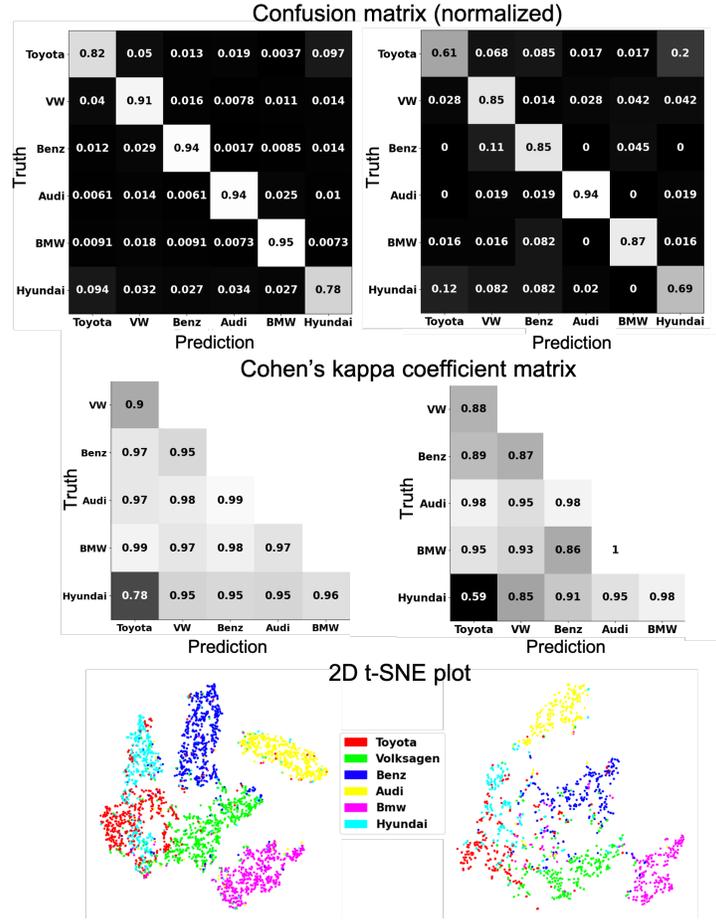

**Figure 12:** BIGNet evaluation on train set (column 1) and test set (column 2) in terms of: Confusion matrix (row 1), Cohen's kappa coefficient matrix (row 2) and 2D t-SNE plot of data's latent vectors (row 3). Both train and test sets show Benz, Audi and BMW having better recognition rate, and that Toyota and Hyundai get confused with each other more often.

### 5.4.2 Performance evaluation

First, confusion matrices are examined. In Figure 12, both train and test sets show a consistent trend of being able to predict Audi, BMW and Benz most correctly. The result also shows that Toyota and Hyundai, while having the least accuracy among the six, also have a relatively higher chance to confuse with each other. Cohen's kappa coefficient matrix is then calculated to validate this finding, as lower values between 0 and 1 indicate more inconsistency between ground truth and model prediction. Dimensional reduction on the last hidden layer using 2D t-SNE also shows evidence that Audi, BMW and Benz are the three distinct clusters in latent space, while Hyundai and Toyota are much more entangled with each other. In other words, Hyundai and Toyota have better segment prototypicality (SP) than other brands, making them harder to differentiate by BIGNet. These findings are consistent with the conclusion of Liu et al. [5], who found that luxurious brands yield higher brand consistency (BC) and that SP has a stronger effect on economy cars than BC.

### 5.4.3 Feature analysis

As mentioned in Section 5.3, the increased data complexity also negatively impacts attempts to visualize the attention of BIGNet using LOFO. LOFO requires to run each image as many times as the number of curves and chunks, leading to huge computational cost. Further, the confidence score change from LOFO often doesn't reflect desirable car features either. The ablation of individual shapes is either too subtle due to the robustness of larger graphs, or is too biased to chunks with large bounding boxes, making the contour always highlighted. This case study, therefore, implements a CAM[17]-inspired algorithm that efficiently runs only one inference per image. In the algorithm, it looks at the contributions of each chunk's latent vector and visualizes the chunks that contribute the most to a correct prediction.

**Table 7:** Most highlighted features by BIGNet.

| brand | number of data observed in test set | 1st most highlighted feature | % | 2nd most highlighted feature | % |
|---|---|---|---|---|---|
| Toyota | 54 | fog lights | 92.6% | logo | 79.6% |
| VW | 70 | logo | 68.6% | grille | 64.3% |
| Benz | 65 | grille | 89.2% | headlights | 76.9% |
| Audi | 63 | grille | 77.8% | headlights | 33.3% |
| Bmw | 65 | headlights | 84.6% | grille | 72.3% |
| Hyundai | 49 | headlights | 69.4% | fog lights | 24.5% |

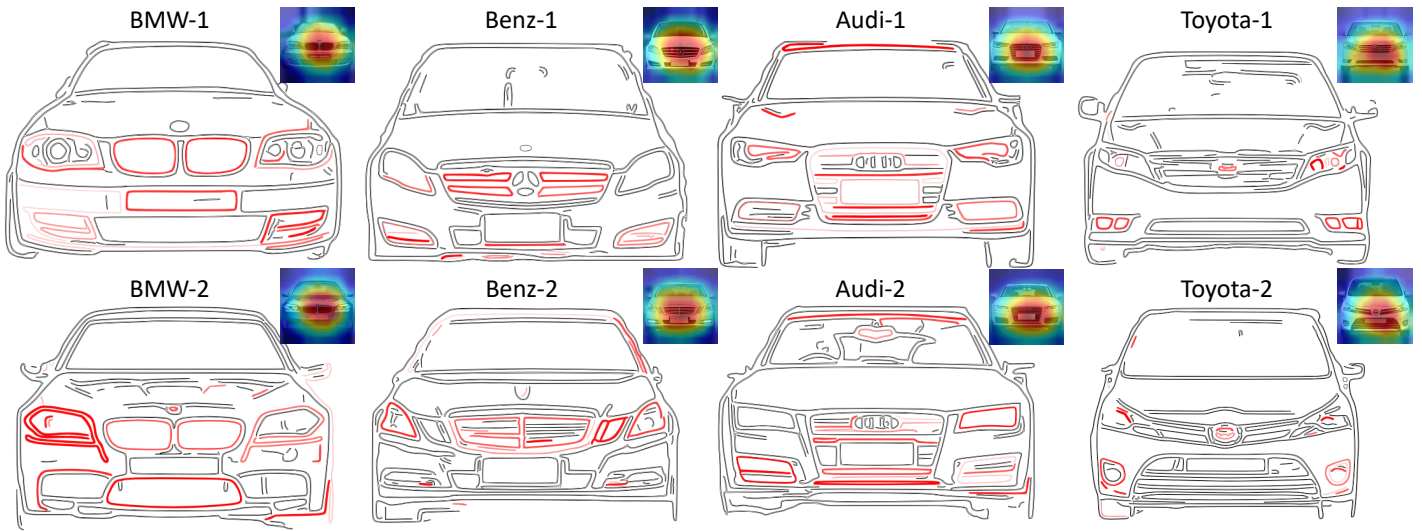

**Figure 13:** CAM-based BIGNet brand-related features' visualization on test set. The Grad-CAM visualizations on CNN are located in the upper right corner of each image. It is obvious that BIGNet captures luxury segments' well-distinguishable car parts including grille, headlights and fog lights, while there are much fewer geometric clues on affordable cars (Toyota) that it has to rely on logo detection.

**Table 8:** Comparison of the four training scenarios. All cases share consistency of having better recognition of luxurious brands and more entanglement between affordable brands.

| brands | has logo | accuracy | Well clustered classes (t-SNE) | Well recognizable (Confusion matrix accuracy) | Badly recognizable (Confusion matrix accuracy) | Entanglement pairs (Cohen's kappa coefficient matrix) |
|---|---|---|---|---|---|---|
| 6 | yes | train: 89.4% test: 80.6% | BMW>Audi>Benz>VW | train: BMW>Audi=Benz>VW test: Audi>BMW>Benz=VW | train: Hyundai<Toyota test: Toyota<Hyundai | Toyota-Hyundai |
| 6 | no | train: 87.1% test: 77.3% | BMW>Audi>Benz>VW | train: BMW>Audi>Benz test: BMW>VW>Benz | train: Toyota<Hyundai test: Hyundai=Toyota | Toyota-Hyundai Toyota-VW |
| 10 | yes | train: 64.9% test: 63.4% | BMW>Audi>Benz | train: BMW>Audi>VW test: Audi>VW>BMW | train: Hyundai<Nissan<KIA test: Ford<Hyundai<Nissan | Toyota-Hyundai Toyota-Nissan Ford-Hyundai Ford-KIA |
| 10 | no | train: 65.8% test: 59.1% | BMW>Audi>Benz | train: BMW>Audi>Benz>VW test: BMW>Audi>VW>Benz | train: Nissan<Ford<Hyundai test: Nissan<Hyundai<Ford | Toyota-Hyundai Toyota-Nissan Ford-Hyundai Ford-KIA |

The visualization results (Figure 13) show BIGNet's consistent attention to certain parts of each brand. Table 7 summarizes the most frequently highlighted attention in test set, and their percentage of being visualized. Among the six brands, the luxury brands (BMW, Benz, Audi) exhibit more explainable and intuitive attention, while all three tending to highlight the curves related to the grille and headlights. This suggests that luxury brands prioritize preserving brand consistency in the same car parts while incorporating different geometries. Finally, ResNet-50 is finetuned as the CNN to compare with BIGNet. Although CNN reaches almost 100% accuracy for both train and test sets, it fails to show the explicit features of a brand using Grad-CAM. BIGNet's results, on the other hand, are not only explainable, but also editable as each curve is parameterized by control points. This makes it a much more useful tool as a surrogate for analyzing brand-related features.

### 5.4.4 Generalizability study

In this section, BIGNet's reliability and consistency are examined by further increasing the classification difficulty, with the effects of logo removal and classifying ten brands (original six brands plus KIA, Chevy, Ford, and Nissan). For each experiment, the accuracy, recognizability ranking, and entanglement pairs are summarized in Table 8. As BMW, Audi, and Benz are frequently getting higher accuracy across all four scenarios, economy cars are also frequently getting lower accuracy and more entanglements, which substantiates the findings in Section 5.4.2. It is also found that adding more brands to the classification problem has a bigger effect than removing the logo, which can be because of all four additional brands are in the economy segments, which have lower brand consistency. It is also notable that logo removal only slightly decreases the accuracy, showing BIGNet's capability of recognizing higher level features. Lastly, for fair comparison, the trained ten-brand classifiers share the same BIGNet architecture that was designed for the six-brand classification task, except for the final linear layer. Therefore, the low accuracy result is very likely because of underfitting, and is possible to be improved by increasing the number of BIGNet's learnable parameters.

## 6 LIMITATIONS AND FUTURE WORK

Although BIGNet's classification accuracy is lower than pixel CNN, there are currently only less than 7000 learnable parameters in BIGNet. This is much fewer than Resnet-50 having 23 million parameters. On one hand, there is a huge space to improve accuracy from hyperparameter tuning and data processing. While the current BIGNet can already demonstrate explicit and explainable visualization results on brand-related features, with increased accuracy, visualization results would be expected to yield even more explainability. On the other, as curve-based images have more condensed information than pixel images, both dataset and AI architecture require smaller storage, which may have applications on lightweight AI design.

This research has shown it possible to extract explicit and editable features by a deep network agent. To help actual designers identify and quantify features in an automatic way, studies are planned to examine how humans can interact and collaborate with such a surrogate system and achieve design objectives. Aside from recognition, there is potential for actively generating or transferring stylized content building upon BIGNet's framework. This may open up a new avenue for future research in data-driven, explainable generative models.

As this research proposes a general stylization workflow, it has wide-ranging applications including image segmentation, engineering design, market positioning and technical appraisement. Aside from classifications to recognize and preserve brand consistency, safety, semantics and ergonomics, with subtle modifications it also has the potential of learning regression problems to recognize or predict contents' labels like year or price. With the strong explainability of learning from curve-based representations, this research opens an avenue of deducing information from curve representations and is expected to outperform pixel base approaches in domains that values interpretability more than accuracy.

## 7 CONCLUSIONS

This research proposes an automatic workflow to analyze and visualize style content explicitly and performs case studies on products' brand classification to recognize and preserve brand consistency. To mimic a human designer's thought process, data is constructed as SVG and is classified using BIGNet, a two-tier spatial GNN. In the phone study, it shows the model able to learn on parametrically synthesized SVG data. By visualizing attention using LOFO, BIGNet demonstrates capability of capturing brand-related features at intra-curve, inter-curve, intra-chunk, inter-chunk levels. Partial dependence plot on model's confidence variation during parameter extrapolation further substantiates that BIGNet learned continuous and meaningful features, including lens' location, height-width ratio, and screen-frame gap. The car study further explores the potential of a fully automated recognition system, and investigates the generalizability of the workflow. With some architecture modifications from the phone study, BIGNet can learn from generic vectorized car images and reach 80.6% test accuracy on a six-brand classification task. During the evaluation process, BMW, Benz, and Audi are found to achieve higher recognition rates compared to other brands. This finding matches the optimized marketing strategy that luxurious cars value brand consistency more than economy cars. CAM visualization further shows that BIGNet has consistent attention on luxury brands' grille and headlights. Finally, comparison of BIGNet with a CNN baseline demonstrates that a curve-based deep learning model produces more interpretable visualizations, while the image format is also more editable. Therefore, BIGNet, as a deep learning model, can identify brand-related features and can be applied to various product categories with distinguishing geometries, enabling humans to finally utilize it as a communicative and explainable style discovery agent, which significantly accelerates the aesthetic design process. Future research will explore a wider range of potential applications in other stylization domains.

### ACKNOWLEDGEMENT

This work was partially funded by the National Science Foundation under grant Award CMMI-2113301.